\pdfoutput=1

\documentclass[11pt]{article}

\usepackage{acl}
\usepackage{times}
\usepackage{latexsym}
\usepackage{amsmath}

\usepackage[T1]{fontenc}

\usepackage[utf8]{inputenc}
\usepackage{amsfonts}
\usepackage{hyperref}
\hypersetup{
    colorlinks=true, 
    linkcolor=blue, 
    filecolor=magenta, 
    urlcolor=cyan, 
}

\usepackage{microtype}

\usepackage{inconsolata}
\usepackage{mathtools}
\title{T3M: Text Guided 3D Human Motion Synthesis from Speech}

\author{
    Wenshuo Peng\textsuperscript{\rm 1},
    Kaipeng Zhang\textsuperscript{\rm 1}\thanks{Corresponding author},
    Sai Qian Zhang\textsuperscript{\rm 2}\thanks{Corresponding author}
    \\
    \textsuperscript{\rm 1}OpenGVLab, Shanghai AI Laboratory\\
    \textsuperscript{\rm 2}New York University\\
    \\
    \texttt{gin2pws@gmail.com, zhangkaipeng@pjlab.org.cn, sai.zhang@nyu.edu}
}

\begin{document}
\maketitle
\begin{abstract}
Speech-driven 3D motion synthesis seeks to create lifelike animations based on human speech, with potential uses in virtual reality, gaming, and the film production. Existing approaches reply solely on speech audio for motion generation, leading to inaccurate and inflexible synthesis results. To mitigate this problem, we introduce a novel text-guided 3D human motion synthesis method, termed \textit{T3M}. Unlike traditional approaches, T3M allows precise control over motion synthesis via textual input, enhancing the degree of diversity and user customization. The experiment results demonstrate that T3M can greatly outperform the state-of-the-art methods in both quantitative metrics and qualitative evaluations. We have publicly released our code at \href{https://github.com/Gloria2tt/T3M.git}{https://github.com/Gloria2tt/T3M.git}
\end{abstract}

\section{Introduction} 
Speech-driven 3D motion synthesis, known as~\textit{speech-to-motion}, is a technique aimed at generating realistic and expressive motion animations from human speech. Despite its promising applications in virtual reality (VR) \cite{wohlgenannt2020virtual}, gaming \cite{ping2013computer}, and film production \cite{ye2022perceiving},  speech-to-motion also encounters significant challenges, involving various modalities and intricate mappings. Speech signals tend to be high-dimensional, noisy, and subject to variability, while motion data often exhibit sparsity, discreteness, and adherence to physical laws. 
Additionally, the connection between speech and motion is not deterministic; instead, it relies on factors such as the environment, emotions, and individual personalities. 

\begin{figure}
\begin{center}
   \includegraphics[width=\linewidth]{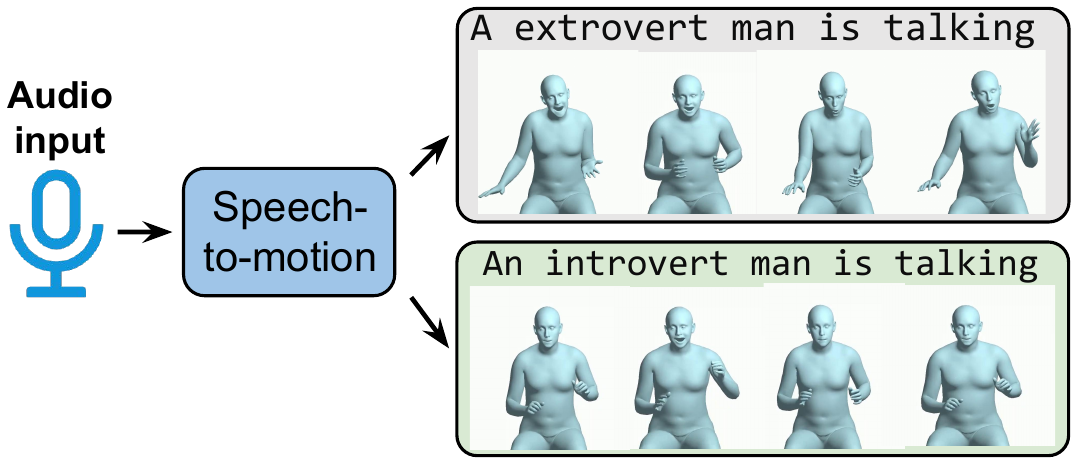}
\end{center}
   \caption{
   Under the same audio input, extrovert and introvert persons will talk in a completely different fashion.
}
\label{fig:fig1}
\end{figure}

Moreover, in traditional speech-to-motion systems, speech audio serves as the sole input for generating various motions for the face, body, and hands. However, this approach may lead to imprecise and undesired motion synthesis due to limitations in the expressive capabilities of the audio signal. Identical audio signals could also stem from entirely unrelated contexts. For instance, as depicted in Figure~\ref{fig:fig1}, when examining the same audio segment, an introverted speaker tends to use minimal body and hand motions compared to an extroverted speaker, who exhibits a more extensive range of movements. Capturing such contextual information solely from audio input proves nearly impossible. This limitation in precise control poses potential difficulties for emerging industries like AI-driven film or animation production, where generated motions may need additional refinement to match user preferences more accurately.

To address this issue, we introduce a novel text-guided 3D human motion synthesis from speech method, termed~\textit{T3M}. The T3M framework enables accurate control of body-hand motion generation via provided text prompts. This improvement is especially valuable for addressing the rigidity often observed in the motions generated due to the relationship between speech and co-speech gesture is one-to-many in nature. Even the same speech in different situation can be result in different motion style. The controllability afforded by T3M facilitates the creation of more nuanced and realistic motion sequences, enhancing overall realism and expressiveness.  

Our T3M contains three major blocks, a VQ-VAE network to generate an intermediate codebook for action-to-action mapping, an audio feature extraction network to extract acoustic information of audio, and a multimodal fusion block to implement audio and text interaction.
Specifically, We train a VQ-VAE network with a two-layer codebook, which contains hand and body information respectively. Considering that human motions are related to the speaker's emotion, intonation, and rhythm, we utilize a pre-trained EnCodec~\cite{defossez2022high} model to extract acoustics features from the original audio. To align the sequenc lenth of audio feature and the stored body-hand motion parameter, we use an audio encoder to downsample the features. Furthermore, we propose a multi-modal fusion encoder structure, which inserts a cross-attention layer to the transformer decoder acoustics architecture for better textual information fusion.  

Another significant challenge arises in the generation of training datasets for T3M. Most existing training datasets for speech-to-motion are in the form of speech-motion pairs, lacking corresponding textual information. One simple approach to address this gap is to utilize a video large language model (VLLM) like Video-Llama~\cite{zhang2023video} for labeling datasets. However, current VLLMs can only provide a coarse-grained description of the video input. 
Additionally, since speech-motion pairs in the training dataset are often extracted from particular segments of lengthy videos, employing VLLM for text generation may lead to highly similar text descriptions being produced across various video clips. To enhance the diversity of textual descriptions within the training dataset, we adopt the video-language contrastive learning framework, VideoCLIP~\cite{xu2021videoclip}. This framework enforces the alignment of video and text in a joint embedding, enabling the processing of video frames and utilizing the resultant video features to replace textual features for T3M training. 

Our research primarily centers on text-guided speech to body and hand motions generation. For 3D face reconstruction, we utilize cutting-edge methods in the field, such as those demonstrated in Emotalk~\cite{emotalk}. Overall, our contributions can be described as follows: 

\begin{itemize} 
    \item We propose a novel speech-to-motion training framework termed~\textit{T3M}, enabling users to achieve better control over the holistic motion generated from audio through the utilization of textual inputs.
    \item  To achieve audio-to-motion generation controlled by text, we align video and text in a joint embedding, utilizing video input for training and text descriptions for inference. This approach notably enhances the diversity of textual input within the training dataset and substantially improves the performance of motion synthesis.
    \item The results show that the proposed T3M framework significantly outperforms existing methods in terms of both quantitative and qualitative evaluations. 
\end{itemize}

\section{Related Work}
\subsection{Motion Generation from Speech}

In recent years, there has been a growing interest in generating human-like motion from speech. One area of research is centered around facial reconstruction, with various studies exploring 2D talking head generation~\cite{mittal2020animating}. These investigations employ image-driven or speech-driven techniques to produce realistic videos of people speaking. 

Extensive research has been conducted in the field of 3D talking heads generation. To make the reconstruction more precise, FaceFormer~\cite{fan2022faceformer} uses a Transformer-based model to obtain contextually relevant audio information and generates continuous facial movements in an autoregressive manner. VOCA~\cite{cudeiro2019capture} uses time convolutions and control parameters to generate realistic character animation from the speech signal and static character mesh. 
MeshTalk~\cite{richard2021meshtalk} places its emphasis on the upper facial generation, an aspect where VOCA falls short. It establishes a categorical latent space for facial animation and effectively separates audio-correlated and audio-uncorrelated motions using cross-modality loss, enabling the generation of audio-uncorrelated actions like blinking and eyebrow movements. 
Another line of research centers on body and hand motion reconstruction. These approaches can be categorized into two groups: rule-based and learning-based methods. Rule-based methods, such as~\cite{kopp2004synthesizing}, involve the mapping of input speech to pre-defined body motion units through manually crafted rules.

The development of learning-based methods for generating body motion, as demonstrated in research like~\cite{ahuja2020no}, has made substantial progress, largely attributed to the availability of openly accessible synchronous speech and body motion datasets~\cite{habibie2021learning}. Very recently, TalkSHOW~\cite{yi2023generating} has introduced a simple encoder-decoder architecture capable of producing holistic 3D mesh motion. 

However, it is important to highlight that despite these advancements, these methods still encounter difficulties in achieving a balance between diverse and controllable motion. Consequently, when applied in real-time scenarios, the generated actions often display repetitiveness and limited adaptability in response to changes in the external condition.

\subsection{Video-text Pre-training}
The aim of video-text pre-training is to utilize the complementary information found in both videos and textual inputs to improve the performance of subsequent tasks.
VideoBERT, as introduced in~\cite{sun2019videobert}, pioneered the exploration of pre-training methods for video-text data pairs. Its primary focus lies in acquiring a unified visual-linguistic representation, and it demonstrates versatility in adapting to a range of tasks, such as action classification and video captioning.

VideoCLIP~\cite{xu2021videoclip} employs a contrastive learning approach to pre-train a unified model for zero-shot understanding of both video and textual inputs, without relying on any labels in downstream tasks. VLM~\cite{xu2021vlm} introduces a simplified, task-agnostic multi-modal pre-training method. This method is capable of handling inputs in the form of either video, text, or a combination of both, and it can be applied to a diverse range of end tasks. 
Recently, there has been a surge in research focused on large language models (LLMs), and some researchers have started incorporating LLMs into this field, yielding promising results. Video-LLaMA~\cite{zhang2023video} bootstraps cross-modal training from the frozen pre-trained visual, audio encoders and the frozen LLMs. This approach utilizes the robust understanding capabilities of large models for tasks such as video understanding and video question answering. In our research, we make use of VideoCLIP to process both video and textual inputs. In particular, during the training phase, we utilize the video encoder of VideoCLIP to convert the video input into latent vector for multimodal learning. During the testing phase, we leverage its text encoder component to enable text-based control over body and hand motion generation.

\section{Method}
In this section, we describe the detailed design of T3M, which can generate holistic body motion, including body poses, hand gestures, and facial expressions, based on provided text descriptions. 
\begin{figure*}
\begin{center}
   \includegraphics[width=1.0\linewidth]{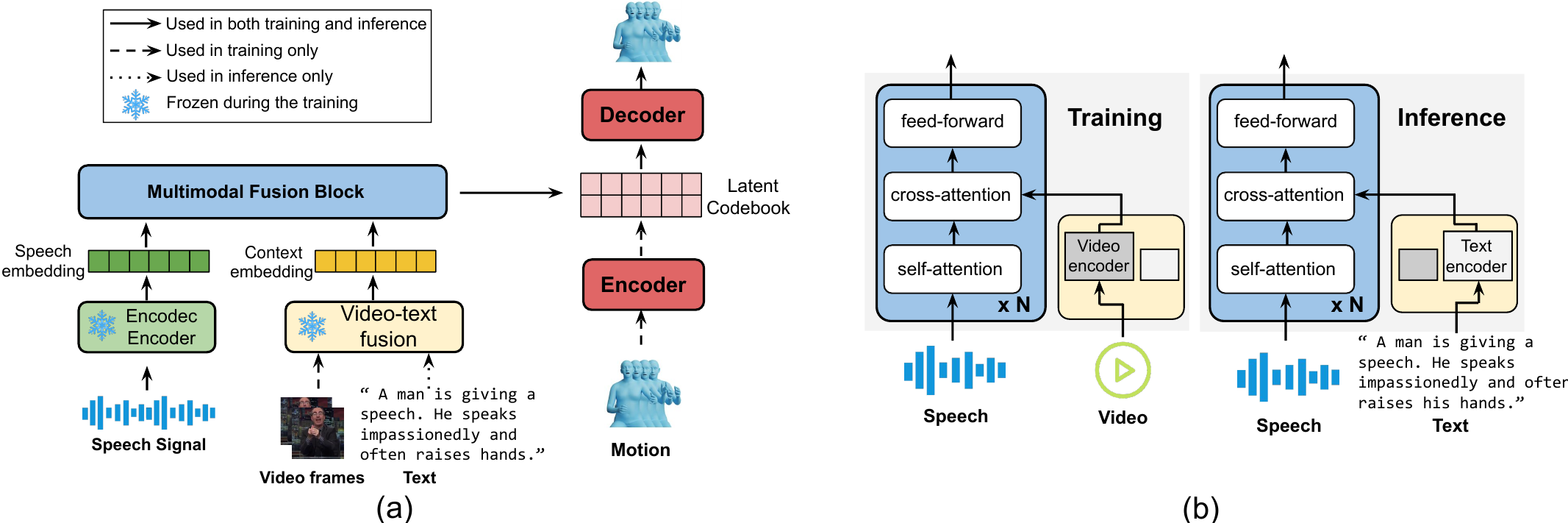}
\end{center}
   \caption{Overview of the proposed T3M. We employ a novel framework for body and hand motion generation. Specifically, T3M first learns a quantized body-hand codebook through a VQ-VAE model. In the training phase, we the pre-trained EnCodec model to extract the speech embedding of the given speech. We employ the pre-trained video encoder from VideoCLIP to obtain the video embedding that corresponds to the provided speech. To facilitate interaction between these two modalities, we utilize a multimodal fusion block. This fusion block is built upon a BERT-based framework, enhanced with a cross-attention layer for effective fusion.
}
\label{fig:overview}
\end{figure*}

\subsection{Preliminary}

We begin by establishing a mathematical formulation for the problem. Specifically, we define a temporal sequence of motion from time $t=1$ to $t=T$ as $A_{1:T}$ further contains three primary components: facial expressions along with the jaw poses denoted as $A^f_{1:T}$, body motions as $A^b_{1:T}$, and hand motions as $A^h_{1:T}$. Each element $a^f_t$ of $A^f_{1:T}$ is defined as $a^f_t = (\theta^{jaw}_t,\zeta_t)$, where $\theta^{jaw}_t$ is the jaw pose and $\zeta_t$ is the facial expression parameter. In the case of $A^b_{1:T}$ and $A^h_{1:T}$, each element is defined as follows: $a^h_t = (\theta^b_t)$, $a^b_t = (\theta^h_t)$, where $\theta^b_t$ and $\theta^h_t$ are the body poses and hand poses, respectively. 
Prior methods~\cite{yi2023generating} primarily produces the holistic motion solely based on the speech input.
In contrast, our objective, when provided with a speech input sequence $S_{1:T}$, is to produce comprehensive motion sequences $A^b_{1:T}$, and $A^h_{1:T}$ by incorporating additional textual context. This context describes the situation and background associated with the speech input, allowing the resultant holistic motions to vary according to both speech and textual inputs. Formally, we express this as:
\begin{equation}
    \hat{A}_{1:T} = F_{T3M}(S_{1:T},B)
\end{equation}
where $B$ is the textual input and  $\hat{A}_{1:T}$ is the output holistic motion generated by T3M and $F_{T3M}$ represents the T3M function. Figure~\ref{fig:overview}(a) depicts the overview of T3M framework, and we will provide a detailed description of each component of T3M in the following sections.

\subsection{Face Generation}

We adopt the approach outlined in TalkSHOW to generate facial expressions and other body parts separately. Given that human facial expressions primarily stem from speech content, we leverage the pre-trained wav2vec 2.0 model~\cite{baevski2020wav2vec} as a semantic encoder to extract semantic representations from the provided speech. These extracted features are then fed into a decoder to reconstruct facial motion. 

The wav2vec 2.0 model consists of three main components: firstly, a stack of convolutional layers that process the raw audio waveform to derive a latent representation; secondly, a group of transformer layers that generate contextualized representations based on the derived latent representation; and finally, a linear projection head that produces the output.

The decoder consists of a Temporal Convolutional Networks (TCNs) with six layers, followed by a fully-connected layer. We employ a similar approach as described in~\cite{yi2023generating} to reconstruct facial motion. However, it is worth noting that within our framework, we can replace the face reconstruction method with other SOTA methods, such as~\cite{emotalk}.

\subsection{Context Features Generation}
\label{sec:prompt-embedding}
As depicted in Figure~\ref{fig:overview}(a), the generation of context feature is a necessary step in T3M training. To create the context embedding, we employ a video-text fusion model designed to generate diverse context features. To achieve this, an intuitive approach involves sending the text description directly to a text encoder, and forward the output context features to the video-text fusion module for further processing. However, this is not feasible for two key reasons. Firstly, it is noteworthy that in many cases, several audio waveform segments within a single video clip exhibit significant textual similarities. This resemblance in textual content results in highly resemble output features across these various audio segments, ultimately leading to a suboptimal overall motion synthesis performance due to lack of training data diversity. In contrast, our approach employs a video encoder to process the video frames corresponding to the audio waveforms, which will capture intricate context features corresponding to each speech segment. These context features are subsequently passed on to the multimodal fusion block for additional processing, as shown in the left part in Figure~\ref{fig:overview}(b). 
Secondly, even though it is feasible to manually design distinct text descriptions with intentional variations for each audio segment, this manual labeling process would be labor-intensive. 

As a result, during training stage of T3M, we choose to utilize the video frames corresponding to the speech input to generate the context feature. During the inference, the text description will be sent to the text encoder for better guiding the holistic motion synthesis (right part of Figure~\ref{fig:overview}(b)).
To enable the precise text-guided motion generation, we adopt the video and text encoders from the VideoCLIP model, which establishes a detailed correlation between video and text through contrastive learning. By mapping video and text embeddings into a common latent feature space, we facilitate seamless modality substitution for text-guided motion synthesis during inference. This approach simplifies the process by eliminating the need for extensive manual labeling of textual descriptions and leveraging existing joint video-text representation models. It builds upon the concept of modality substitution, which has been successfully employed in other contexts, such as MusicLM~\cite{2023musiclm} with audio and text.

\subsection{Body and hand Motion Generation}
\paragraph{Audio Feature Encoder} We obtain the speech embedding using EnCodec~\cite{defossez2022high}, a state-of-the-art neural audio codec pre-trained model capable of extracting audio features from the provided speech. It contains a total of eight layers of codebooks, each layer stores different audio information. Given that the audio token sequence stored in the 8-layer codebook is overly lengthy, we opt not to employ these audio tokens as the initial input. Instead, we employ the decoder of the codebook to generate the audio features, which serve as our speech embeddings. 
Next, we employ a compression model based on a convolutional neural network architecture to transform these features, aligning them with the sequence length of the motion embeddings.
Finally, we further insert an MLP at the end of the compression model to map the dimensions to 768, where 768 is the dimension dim of the context feature.
Using our audio feature encoder, for a speech segment lasting $k$ seconds, we obtain an speech embedding with a dimension of $e^a \in \mathbb{R}^{L_{seq}\times768}$, where $L_{seq}=k\times{fps}$ represents the sequence length, where $fps$ is the frames per second rate of the motion data. This rate determines the number of motion frames that correspond to each second of speech, thereby aligning the temporal resolution of the audio and motion data.

\paragraph{Latent Codebook Design}

It is challenging to directly produce the body-hand motion sequence for a given speech sequence because the input and output belong to two distinct modalities. To mitigate this problem, we utilize the VQ-VAE~\cite{van2017neural} model to create a latent codebook for both body motion and hand motion. 

Consequently, we obtain two distinct finite codebooks: $Z_b = \{z_{b_i}\}_{i=1}^{|Z_b|}$ for body motions and $Z_h = \{z_{h_j}\}_{j=1}^{|Z_h|}$ for hand motions, where $z_{b_i}, z_{h_j} \in \mathbb{R}^{d_z}$ and $d_z$ denotes the length of each codebook element. This approach yields $|Z_b| \times |Z_h|$ different body-hand pose code pairs $(z_{b_i}, z_{h_j})$, significantly expanding the range of motion diversity.

\paragraph{Multimodal Fusion Block Design}

The Multimodal Fusion Block is a transformer-decoder based model that incorporates an cross-attention layer between the feedforward layer and the self-attention layer, as depicted in Figure~\ref{fig:overview}(b). Its purpose is to produce latent codebook tokens from the provided speech features and context features, which serve as input for the VQ-VAE decoder.

As described in Section~\ref{sec:prompt-embedding}, during the training phase, we substitute the text input with the video frames corresponding to the speech. We encode these video frames by ViCLIP video encoder into the context feature space, which is shared with the text captions. Thus,  for the video $x$, its corresponding feature can be derived as follows: 
\begin{equation}
    e^{v} = F_v(x)
\end{equation}
where $e^{v} \in \mathbb{R}^{1\times512}$ is the feature vector for the video $x$ and $F_v$ is the video encoder function. $e^{v}$ will be used as the context features during the inference operation for conditional body and hand motion generation.

For the speech embedding $e^{a}$ and the context features $e^{v}$, the multimodal fusion block layer then combines them through standard cross-attention. It is important to highlight that the cross-attention layer between the speech features and the context features offers two significant advantages. Firstly, our model integrates context features during training, enabling the generation of distinct body-hand motions based on varying input text during the inference stage. Secondly, as this context feature is incorporated during training, the reconstructed motion exhibits higher quality and a greater level of alignment.

\subsection{Loss Function}
As illustrated in Figure~\ref{fig:overview}(a), the training process of T3M involves three main stages. First, the facial image generator is trained to convert audio signals into facial expressions. Semantic features are extracted using a pre-trained wave2vec encoder, and the decoder is trained to minimize the Mean Square Error (MSE) loss between the ground truth facial output and the decoder output.

Second, the VQ-VAE model is trained to map body and hand motions into a latent space, resulting in a codebook $C \in \mathbb{R}^{d_z \times 2}$. Formally, we have
\begin{align*}
    \mathcal{L}_{V Q} &= \mathcal{L}_{\text{rec}}(A, \widehat{A}) + \alpha \left\| \text{sg}[z_e(A)] - Z_Q({A)} \right\| \\
    &+ \lambda \left\| z_e(A) - \text{sg}[z_q(A)] \right\|
\end{align*}
 where $\mathcal{L}_{\text{rec}}$ is the mean squared error reconstruction loss, $sg[.]$ is the stop gradient operation,  $z_e$ is the output of the VQ-VAE encoder and $z_q$ is the quantization function. $\alpha$ and $\lambda$ are two weight coefficients to reflect the importance of each component.

 Lastly, we train our multimodal fusion block to generate the discrete token of the codebook from the speech and video through cross-entropy loss.

\section{Experiment}
We first describe experiment setup in Section~\ref{sec:setup}, and provide quantitative and qualitative evaluate in Section~\ref{sec:quantitative} and Section~\ref{sec:qualitative}. We then evaluate T3M over more data examples in Section~\ref{sec:other_examples}, and present the ablation study in Section~\ref{sec:ablation}. We include a video demo in the supplementary materials (data).
\subsection{Experiment Setup}
\label{sec:setup}
\paragraph{Dataset}
In our research, we use the SHOW dataset~\cite{yi2023generating}, a high-quality audiovisual 
dataset which consists of expressive 3D body meshes at 30fps, with a synchronized audio at a 22K sample rate. The 3D body meshes are reconstructed from in-the-wild monocular videos and are used as our pseudo ground truth (p-GT) in speech-to-motion generation. For a given video clip of $T$ frames, the p-GT comprises parameters of a shared body shape $\beta \in \mathbb{R}^{300}$, poses $\theta_t \in \mathbb{R}^{156T}_{t=1}$, a shared camera pose $\theta_c \in \mathbb{R}^3$, a translation $\epsilon \in \mathbb{R}^3$ and facial expressions $\psi_t \in \mathbb{R}^{100T}_{t=1}$. Here, the pose $\theta_t$ includes the jaw pose $\theta_{\text{jaw}_t} \in \mathbb{R}^3$, the body pose $\theta_{b_t} \in \mathbb{R}^{63}$, and the hand pose $\theta_{h_t} \in \mathbb{R}^{90}$.

\paragraph{Compared Baseline}
We compare T3M with TalkSHOW, the first research work on holistic 3D human motion generation using speech. We also evaluate the authenticity and diversity of the resultant motion synthesis by comparing various baselines, including Audio Encoder-Decoder~\cite{ginosar2019learning}, Audio VAE~\cite{yi2023generating}, and Audio+Motion VAE~\cite{yi2023generating}.

\paragraph{Metrics}
We used the following methods to measure the quality of the generated holistic motion. Firstly, we calculate the~\textit{Reality Score (RS)} of the generated body and hand motions by employing a binary classifier, as per the methodology outlined in~\cite{aliakbarian2020stochastic}. The classifier is trained to distinguish between authentic and synthetic samples, and RS is computed from its predictions, serving as a metric for assessing the realism of the generated motions. Secondly, we compute the Beat Consistency Score (BCS)~\cite{taming} of the resultant motions to evaluate the motion-speech beat correlation (i.e., time consistency).

\begin{table}
\centering
\begin{tabular}{lcc}
\hline
\textbf{Method} & \textbf{RS} & \textbf{BCS} \\ \hline
Habbie et al. & 0.146 & - \\ 
Audio Encoder-Decoder & 0.214 & - \\ 
Audio VAE & 0.182 & - \\ 
Audio+Motion VAE & 0.240 & - \\ 
TalkSHOW & 0.414 &0.8130 \\ 
T3M(video prompt) & \textbf{0.483}&\textbf{0.8586} \\
T3M(random prompt)& 0.364 &0.8398 \\
\hline
\end{tabular}
\caption{Evaluation results on several methods. For convenience, we use video prompt and random prompt to test our T3M. - means the results are not available. We focus on the comparison with TalkSHOW.}
\label{tab2}
\end{table}

\subsection{Quantitative Evaluation}
\label{sec:quantitative}

Our experiment results are presented in Table~\ref{tab2}. When using T3M,  we utilize two distinct prompt types for generating the context features. The initial type replicates the training stage, utilizing a video prompt. In contrast, the second type entails the generation of a random vector with a mean of $-0.04$ and a variance of $0.12$, which is utilized as the context features. In this setup, T3M generates synthetic motions solely based on the speech input.

Based on the data presented in Table~\ref{tab2}, it is evident that our T3M, when using a video prompt, demonstrates superior performance in terms of both RS and BCS indicators. Furthermore, we note that employing a random prompt yields a slightly lower RS score compared to TalkSHOW; however, it outperforms TalkSHOW in terms of BCS, which demonstrate that the generated motions by our T3M are more consistent with the audio,

\subsection{Qualitative Evaluation}
\label{sec:qualitative}

\begin{figure*}
\begin{center}
   \includegraphics[width=\linewidth]{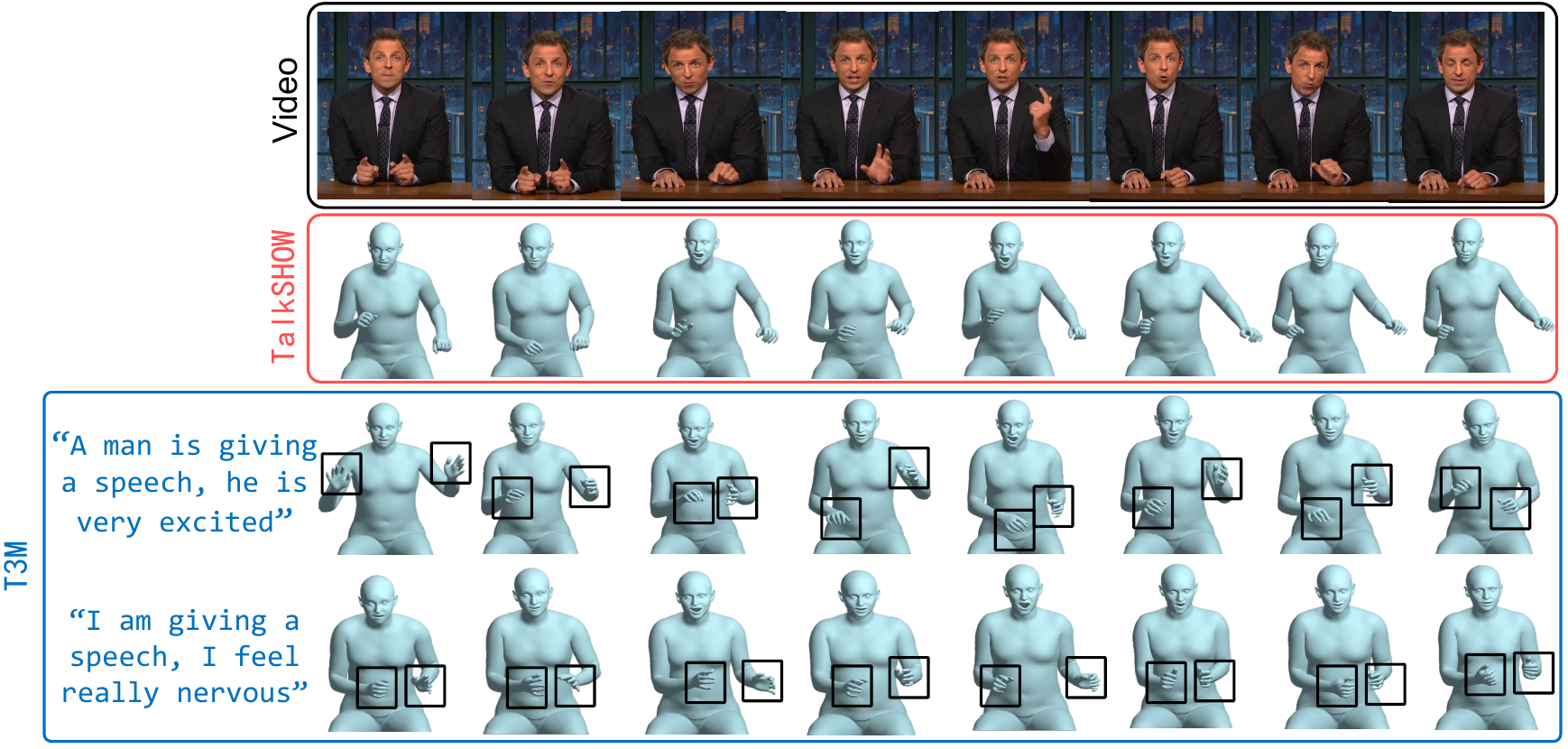}
\end{center}
   \caption{Visualization of 3D holistic motions generated by TalkSHOW and T3M. For T3M, three different text prompts are provided and the positions of the hand are highlighted with black boxes. We notice that the hand motions are closely aligned with the input text desription in T3M.
}
\label{fig:ana}
\end{figure*}
\paragraph{Visualization Results} 
To demonstrate the impact of textual input over the resultant motion synthesis, we utilized two text prompts with opposing semantic meanings, along with a randomly generated embedding as our prompt input. Specifically, for the text prompts, we use ``A man is giving a speech, he is very excited'' and ``I am giving a speech, I feel really nervous'', which has totally opposite semantic meanings. Additionally, we also compare the resultant holistic motions with TalkSHOW, the visualization examples are depicted in Figure \ref{fig:ana}.

As depicted in Figure \ref{fig:ana}, it is evident that the motions generated by TalkSHOW appear to lack diversity. The motions of both hands change independently and do not correspond to the speaker's emotions and intonation, resulting in a unnatural and unrealistic appearance. For T3M, When using the random prompt, we notice that the hand motions closely resemble those of the original video. Additionally, the motions of both hands exhibit corresponding interactions and a higher coordination. 

When using the prompt input, ``A man is giving a speech, he is very excited.'', we observe a notable increase in the range of hand motion changes. Additionally, there are noticeable upward and downward movements of the hands. These motions align closely with our textual description, reflecting the speaker's highly excited state during the speech. By comparison, for text description ``I am giving a speech, I feel really nervous''. We notice that the generated motions distinctly portray signs of nervousness. The hand movements are very restricted, and there are noticeable trembling or jittery motions, effectively capturing the heightened nervous state of the speaker. Overall, the experimental results show that with the introduction of textual input, T3M can achieve controllable motion generation with much higher degree of diversity.

\paragraph{User Study}

\begin{table}
\centering
\begin{tabular}{lcc}
\hline
\textbf{Method} & \textbf{hands and body} & \textbf{holistic}  \\ \hline  
TalkSHOW & 3.43 & 3.36 \\  
T3M (random) & 3.25 & 3.08 \\  
T3M (video) & 3.86 & 3.95 \\ 
\hline
\end{tabular}
\caption{User study results (higher scores indicating better quality). We use the video prompt and the random prompt to evaluate the quality of our generated motions.}
\label{tab3}
\end{table}

To offer a more comprehensive evaluation of T3M, we have devised a thorough user study questionnaire. Following the methodology employed in TalkSHOW, we randomly selected 40 videos from four different speakers in the SHOW dataset, with each video having a duration of 10 seconds. We have invited 12 participants to participate in the evaluation process. 
Each participant will give a score ranging from 1 to 5 to rate the video in terms of the generated motions. We use random prompt and video prompt for our T3M model. Subsequently, we compute the average scores and document the results in Table~\ref{tab3}.

Table \ref{tab3} reveals that our T3M, when using the video prompt, attains the highest scores. Furthermore, it is evident from the table that utilizing a random prompt yields only slightly lower scores compared to the TalkSHOW method.

\subsection{Other Examples}
\label{sec:other_examples}
\begin{figure}
\begin{center}
   \includegraphics[width=1.0\linewidth]{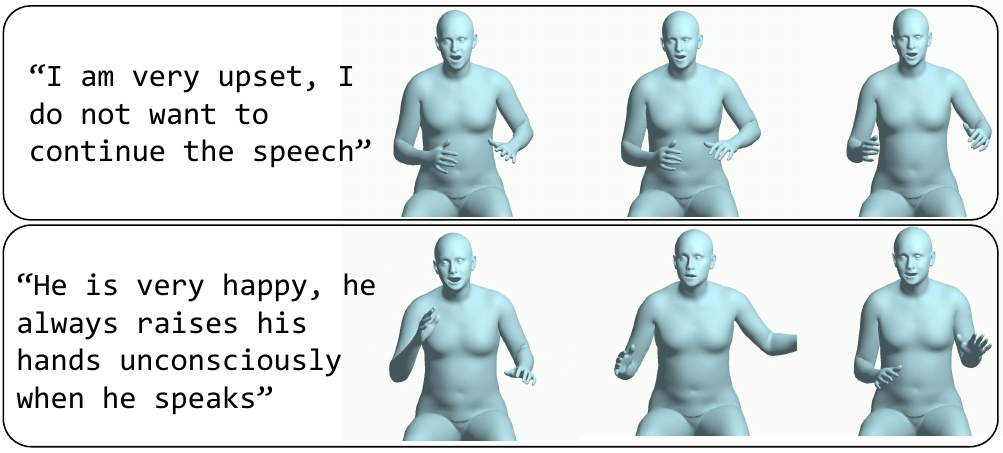}
\end{center}
   \caption{ Experiments on unseen speech. We use two different text input to control the motion generation.}
\label{fig:other}
\end{figure}


In order to better verify the effect of T3M, we evaluate samples that are not contained in the SHOW dataset. We employ an audio clip of French as our speech input, utilizing two textual descriptions as prompts to enhance our evaluation. One text expresses strong negative emotions:  ``I am very upset, I do not want to continue the speech'', while the other conveys positive emotions: ``He is very happy, he always raises his hands unconsciously when he speaks''. The results are shown in Figure~\ref{fig:other}.

We have noticed that when dealing with speech not present in the SHOW dataset, T3M is still capable of producing distinct actions in response to the input text. Specifically, when using `` I am very upset, I do not want to continue the speech'', there is a lack of noticeable alterations in the accompanying hand movements to convey the speaker's upsetness.  When using ``He is very happy, he always raises his hands unconsciously when he speaks'', the range of hand movements of the speaker increased significantly, including a conspicuous pattern of raising the hands.  These findings demonstrate that our approach successfully accomplishes motion generation even in zero-shot scenarios.

\begin{table}
\begin{center}
\begin{tabular}{lcc}
\hline
\textbf{Method} & \textbf{BCS} & \textbf{Motion Score}  \\ \hline
T3M (MFCC) & 0.8050 & 3.54 \\
T3M (EnCodec) & 0.8586 & 3.82 \\ 
\hline
\end{tabular}
\caption{Ablation results between Mel Frequency Cepstral Coefficients (MFCC) and EnCodec. For the BCS indicator, using EnCodec achieves $+0.0536$ test scores. For the user study score of the motion score indicator, we randomly selected a total number of 20 generated videos and invited 5 people to rate them.}
\label{tab4}
\end{center}
\end{table}

\subsection{Ablation Study}
\label{sec:ablation}

We conduct an ablation study to examine the contribution of each component in T3M model. 
\paragraph{Effect of EnCodec} We replace the EnCodec with Mel Frequency Cepstral Coefficients (MFCC)~\cite{zheng2001comparison} and use the BCS and user study score (USS) to measure the effectiveness of generated motions. We report the results in Table~\ref{tab4}. Comparing MFCC with EnCodec, we observe a noticeable performance improvement when utilizing EnCodec. Specifically, an increase of 0.0536 in BCS is observed with the usage of EnCodec. A user study was conducted to evaluate the motion score. A total of 20 samples are randomly selected for evaluation. Five individuals were invited to rate the generated videos, and a higher score indicates better performance. From Table~\ref{tab4}, we also observe T3M with EnCodec achieves a better performance over MFCC.

\paragraph{Impact of Context Features} We aim to investigate the impact of context feature over the synthesis effect. Particularly, we use four different types of embeddings to encode context: random prompt, text prompt, video prompt, zero prompt. For the text prompt, we use ``I am giving a speech, I feel really excite''. In contrast, for the zero prompt, we employ a context feature vector consisting entirely of zeros. We invite five individuals to rate ten videos which are generated from ten randomly selected speech samples. We present the USS in Table~\ref{tab5}. We observe that text prompt and the video prompt both achieve better performance over random prompt and zero prompt.
\begin{table}
\begin{center}
\begin{tabular}{lc}
\hline
\textbf{Method} & \textbf{Motion Score}  \\ \hline
T3M (random) & 3.12 \\
T3M (zero) & 2.52 \\
T3M (text) &  3.69 \\
T3M (video) &  3.85 \\
\hline
\end{tabular}
\caption{Ablation results to evaluate different context embeddings. "Zero" means using an all-0 vector as the context. We use the user study results to evaluate the generated motions.}
\label{tab5}
\end{center}
\end{table}

\section{Conclusion}
In this paper, we proposed T3M, a novel text-guided 3D human motion synthesis method from speech. T3M can generate realistic and expressive holistic motions by leveraging both speech and textual inputs. We use a pre-trained EnCodec model to extract audio features from speech and a multi-modal fusion model to fuse the audio and text features. To enhance the text diversity during training, we employed VideoCLIP, a video-language contrastive learning framework, to process the video frames and use the output video features to replace the textual features. By training on the SHOW dataset, a 3D holistic dataset, T3M enables users to precisely control the holistic motion generated from speech by utilizing textual inputs.

\section{Limitation}
When considering future enhancements, it is possible to achieve even better performance with T3M by incorporating more advanced text and video encoders from a pretrained multimodal model that surpasses the capabilities of VideoCLIP.
To the best of our knowledge, the SHOW dataset currently stands as the sole dataset in the field of speech-driven 3D motion synthesis, albeit it covers a relatively limited range of scenes. We believe that enhancing the performance of T3M could be achieved by training it on more extensive datasets that involves a wider variety of scenarios and contexts.


\bibliography{acl_latex}
\appendix

\label{sec:appendix}
\section{Implementation Details}
\subsection{Quantization
Function of VQ-VAE}
Given the input body motions $A_b^{1:T} \in \mathbb{R}^{63 \times T}$ and hand motions $A_h^{1:T} \in \mathbb{R}^{90 \times T}$, the encoding process begins by mapping them to feature sequences. Specifically, we obtain $E_{b1:\tau} = (e_{b1}, \ldots, e_{b\tau}) \in \mathbb{R}^{64 \times \tau}$ and $E_{h1:\tau} = (e_{h1}, \ldots, e_{h\tau}) \in \mathbb{R}^{64 \times \tau}$, where $\tau = T \cdot C$ and $C$ represents the temporal window size. In our experiment, we set $C = 4$ to strike a balance between the speed of inference and the quality of the feature embeddings.

For the quantization, we have
\begin{align*}
    z_{b_t} &= \arg \min_{z_{b_k} \in Z_b} \| e_{b_t} - z_{b_k} \| \in \mathbb{R}^{64}, \\
    z_{h_t} &= \arg \min_{z_{h_k} \in Z_h} \| e_{h_t} - z_{h_k} \| \in \mathbb{R}^{64}.
\end{align*}

Here, $z_{b_t}$ and $z_{h_t}$ represent the quantized embeddings for body and hand motions at time $t$, and $Z_b$ and $Z_h$ denote the codebooks associated with body and hand motions, respectively.

\subsection{Training Details}
\paragraph{Face Generator}For the head reconstruction, We adopt SGD with momentum and a learning rate of  0.001 as the optimizer. The face generator is trained with batchsize of 1 for 100 epochs, in which each batch contains a full-length audio and corresponding facial motions.

\paragraph{VQ-VAE}For the VQ-VAE training, the VQ-VAE processes input consisting of either body or hand motions. Each VQ-VAE encoder is constructed with three residual layers, incorporating temporal convolution layers with a kernel size, stride, and padding of 3, 1, and 1, respectively. Batch normalization \cite{ioffe2015batch} and a Leaky ReLU activation function\cite{maas2013rectifier} follow each convolution layer. An additional temporal convolution layer with a kernel size, stride, and padding of 4, 2, and 1, respectively, is interleaved after every residual layer, except the last, to maintain a temporal window size (\( C \)) equal to 4. A fully connected layer is added atop the encoder to reduce dimensions before quantization. The decoder mirrors the structure of the encoder. For optimization, Adam is employed with \(\beta_1 = 0.9\), \(\beta_2 = 0.999\), and a learning rate of 0.0001. The weight (\(\beta\)) for the commitment loss is set to 0.25. Training of the VQ-VAEs is conducted with a batch size of 128 and a sequence length of 88 frames for 100 epochs.

\paragraph{Multimodal Fusion} For the given speech embedding from EnCodec and context embedding from VideoCLIP, we first use a compression model to downsample the speech embedding. The compression model is a superposition of 3 one-dimensional convolutions and residual layer \cite{he2016deep} and use ReLU activation function. 
For the multimodal fusion block, we set the number of attention heads to be 8. We set the total number of hidden layer to be 6, respectively. For optimization, we use Adam with a learning rate of 1e-4 and we use the cosin warmup schedule. 

\end{document}